# Survey on the Convergence of Machine Learning and Blockchain

Shengwen Ding[1], Chenhui Hu[2]

`kamisama.ding@gmail.com, chenhui.hu@gmail.com`

**Abstract.** Machine learning (ML) has been pervasively researched nowadays and it has been applied in many aspects of real life. Nevertheless, issues of model and data still accompany the development of ML. For instance, training of traditional ML models is limited to the access of data sets, which are generally proprietary; published ML models may soon be out of date without an update of new data and continuous training; malicious data contributors may upload wrongly labeled data that leads to undesirable training results; and the abuse of private data and data leakage also exit. With the utilization of blockchain, an emerging and swiftly developing technology, these problems can be efficiently solved. In this paper, we survey the convergence of collaborative ML and blockchain. Different ways of the combination of these two technologies are investigated and their fields of application are examined. Discussion on the limitations of current research and their future directions are also included.

## 1　Introduction

Machine learning (ML) has been pervasively researched nowadays and has been applied in many aspects of real life. Tens of thousands of data generated by end-users every day can be used for ML model training, and these trained models can be utilized to solve many problems in industry or daily life. Nevertheless, issues of model and data still accompany the development of ML. For instance, current training on ML models generally requires large amounts of data, which are often not available in practice [1] or limited to the high cost of collection [2]; filtering out "bad data" is a constant battle with spammers or malicious contributors, who can submit low-effort or nonsensical data and still receive rewards for their work [3]; it is also hard to generalize ML models to reflect future due to out-of-date training [4]; concerns about data privacy and data leakage still exist [5] in fields such as Industrial Internet of

Things [6], etc. In this paper, we mainly focus on the solutions to the following issues:

**Limited access to data sets.** The training of ML models requires large amounts of data in general. However, most of the data sets that match the model are proprietary, which means the ones who want to train ML models on certain problems may have limited access to the necessary data sets or they have to spend a very high cost on data collection, thus restricting the training of the models. On the other hand, data sets holders trying to solve certain issues may find it difficult to build or seek ML models that meet their expectations. So, it appears that a marketplace that encourages data and ML models exchanging may be useful and marketable.

**Out of date models.** When a ML model is trained and published, it will soon become out of date without continuous training on a new data set. In other words, an un-updated ML model may fail to make accurate results and reflect future trends when new test data sets become available, thus lacking model generalization.

**Malicious contributors and bad data.** Even if the updatable data is available to the ML model, the data contributors may not be trustworthy. Spammers or malicious data providers may upload wrongly labeled data to the model that leads to undesirable training results. In this case, the updated ML model may perform badly if not filtering out these bad data.

**Abuse of private data and data leakage.** In traditional ML model training, it is difficult to track the users and their utility of certain data. Therefore, data contributors may be fear the abuse of their private data or data leakage and be reluctant to share their data sets. Since privacy is a key prerequisite, this paper also gives feasible schemes to ensure this.

Obviously, the above deficiencies hinder the development of ML, hence it is essential to improve the traditional ML model training. With the utilization of blockchain, these problems can be efficiently solved. Blockchain is a peer-to-peer network used to record a public history of transactions without relying on a trusted third-party [7]. It is an emerging and swiftly developing technology and it has the advantages of decentralization, privacy protection, immutability, and traceability. A further explanation of blockchain will be given in Section 2.

An increasing number of researchers have been paying attention to the convergence of ML and blockchain and solving problems that exist in traditional ML model training. As is mentioned above, training of traditional ML models is limited to access to data sets, which are generally proprietary or cost highly. To address this concern, a framework of Sharing Updatable Model (SUM) on blockchain [8] can be used to incentivize data contributors to the ML model and help continuous training. Besides, its incentive mechanism can also reduce the negative influence of malicious contributors or spammers and their bad data by detaining their deposits of them. To ensure data privacy and avoid data abuse, Federated learning (FL) [9], a well-studied ML framework, has been utilized to protect the privacy of data while allowing collaborative ML model training. Although FL enables contributors to share the federated models learned over decentralized parties, it may still face the problems of malicious servers and unbelievable participants who may behave incorrectly in ingredient collecting or parameter updating [10]. And so blockchain empowered FL system [11] and blockchain-based incentive systems such as DeepChain [10] have come into use.

Besides data sharing, there are also researches on the marketplace of ML model sharing, which not only helps cooperate the participants with trained ML models and those with data but also helps update the outmoded ML models. DanKu1 contract [12] is one classic protocol built on top of Ethereum blockchain [13] for evaluating and exchanging ML models. A more completed marketplace for both ML model and data sharing is conducted on the Architecture of DInEMMo [14], a convergence of decentralized AI and blockchain. In these marketplaces, transactional privacy is also a key issue, so a decentralized smart contract system named Hawk [15] can be taken into consideration.

Consequently, it is clear that when ML is combined with blockchain, the latter can help improve the quality of the data to train more performant ML models. By encouraging data sharing and model sharing, researchers will not have to worry about the limited access to the latest data and making models lack generalization, and malicious data that is unfavorable to model training can be screened out by the validation mechanism built in the smart contract. Besides, due to the encryption mechanism provided by the blockchain, data providers will not have to fear that their private data will be leaked or abused, avoiding the hidden risks of information security. The convergence of ML and blockchain will have a far-reaching significance since it can be used for incentivizing the creation of better ML models and making them more accessible to companies in various fields such as healthcare, financial service, supply chain, etc. [16]

In the course of our survey, the papers and experiments we select are mainly related to the methods and strategies for combining ML and blockchain, in which state-of-the-art articles of this type are highlighted in this paper. The most representative works are classified into two categories and their key assumptions, strategy architectures, and main results are described in a detailed manner.

The rest of this article is organized as follows. In Section 2, we introduce the background of ML and blockchain; in Section 3, the applications of the convergence of ML and blockchain and their innovations are described in detail; The limitations and future research directions of these applications are listed in Section 4, and the conclusion of our work is in Section 5.

## 2 Background

In this section, we respectively introduce the brief background of ML and blockchain.

### 2.1 Brief Introduction to Machine Learning

ML has been widely researched so far. A variety of ML models are applied in different fields, such as lip reading [17], speech recognition [18], image classification [19], etc. to solve various types of problems.

Classical algorithms of the ML model include support vector machines (SVM), decision tree, random forests, [20], etc. One of the use cases is the classification problem, which is the ability to identify and classify discrete variables such as texts, images, or documents into different classes. Another use case of these three algorithms is the regression problems which predict continuous variables like forecasting temperature. SVM is a supervised algorithm that finds an optimal separation line (hyperplane) to separate data points into different classes. To understand the random forests, we need to know the decision tree first. Like its name, the decision tree is a tree-like model with roots and branches and identifies possible results by using control statements. Random forest is a supervised algorithm that combines multiple decision trees to work as a whole. Its combination of outputs outperforms any individual decision tree model.

---

[1] DanKu Contract is named after the combination of the family names of its developers, K. **Dan**iel and A. B. **Ku**rtulmus.

Another typical ML model is called Neural Network (NN), which can learn more complicated patterns from its input data and solve complex classification and regression problems [20], compared with classical models. For example, in the case of classifying a huge amount of text data (with a great number of patterns), NN may perform better than SVM. NN is made out of nodes, biases, and weighted edges and can represent virtually any function [21]. A basic schema of NN is shown in **Fig. 1**. There are three types of layers in a NN model: input layer, hidden layer, and output layer. The nodes of an input layer are the input data used to train the model. The biases and weighted edges (green arrows in **Fig. 1**.) of the model are adjusted to increase model performance. The hidden layer acts as a black box to figure out the relationship between the input data and the output labels. The output layer is the result of the trained model. After the training step, the accuracy of the model is validated by using a set of testing data. The model accuracy is a key value that represents the performance of a trained model and is dependent on the amount of available training data. Therefore, many researchers are encouraging the promotion of data sharing. Once being trained, the NN model can be applied to perform on new datasets.

Since ML is purely software and its training doesn't require interacting physically, it is a natural choice for the researchers to combine ML with blockchain for coordination between users with data and ML models.

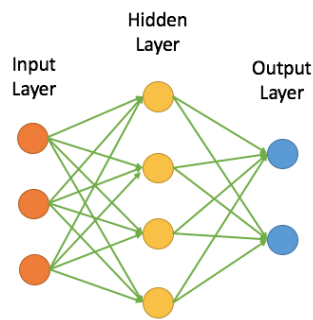

**Fig. 1.** A schema of a single layer fully connected Neural Network.

## 2.2 Brief Introduction to Blockchain

When bitcoin, a peer-to-peer electronic cash system, was firstly introduced by Satoshi Nakamoto in January 2008 [7], the blockchain technology that maintains the transactions ledger of bitcoin was also starting to receive public attention. Satoshi's blockchain was the first decentralized solution that allows for public agreement (consensus mechanism) on the order of transactions without going through a third party. A basic schema of blockchain is shown in **Fig. 2**. Blockchain gets the name from its way of storing data. The system of blockchain creates a persistent and ever-growing chain of "blocks" that contain timestamps, nonces, hashcodes, previous block's hashcodes, and lists of transactions. The previous hashcode acts as the "chain" that links each "block". Since every "block" refers to the hashcode of the previous "block", it is impossible to alter the information of one single "block".

As an emerging technology, blockchain functions with the following properties:

**Decentralization.** Blockchain is a decentralized peer-to-peer network (ledger) that eliminates the reliance on a third-party middleman, for example, the payment apps like PayPal and Alipay. This helps improve process efficiency and reduces transaction costs.

**Privacy Protection.** While information needs to be verified through a consensus process before it is added to the blockchain, the data itself is converted into a series of letters and numbers through a hashcode. Only the participants with a key can decipher this information in the blockchain network.

**Immutability.** All participants with the key can view the information in the blockchain, but they can hardly modify the data recorded. This is helpful to reduce the risk of fraud and build a trustworthy system.

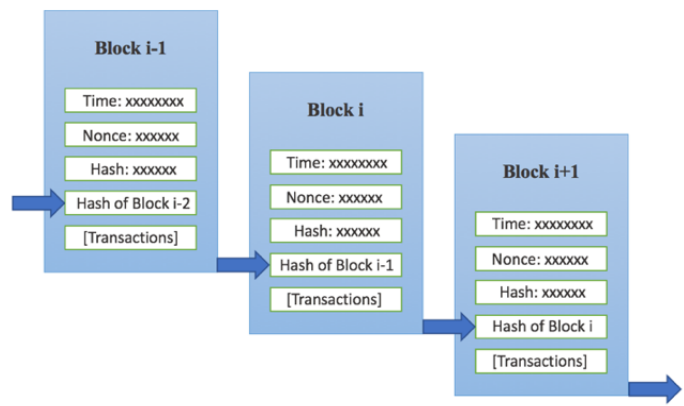

**Fig. 2.** A schema of a blockchain: Each "block" in the blockchain contains a timestamp, a nonce (number only used once), a hashcode, the hashcode from the previous "block", and a transaction list. The timestamp records the time when data are hashed into the blockchain. A nonce is a number that the miners of the blockchain are solving for before adding it to a hashed "block" in the process of cryptocurrency mining. The blockchain will meet the difficulty level restrictions when rehashed. The hashcode is a sequence of numbers and letters that ensures encryption. The hashcode from the previous block acts as the "chain" between the "blocks". Transaction lists are the data recorded by each "block".

**Traceability.** Since the information recorded in the blockchain is immutable, it is ideal for tracing and tracking the use of data. In this case, data abuse and data leakage can be decreased.

Ethereum is one of the most successful and widely applied blockchains with a built-in Turing-complete programming language, allowing anyone to write "contracts" (called smart contracts) and decentralized applications simply by writing up the logic in a few lines of code [13].

The nature of blockchain has a far-reaching potential to offer dramatic advantages to businesses and daily life. So far, it has been widely used in many areas such as financial service [16], supply chain [16], healthcare [22] [23], smart city [24] [25], etc.

## 3 Applications and Innovations

In this survey, we mainly investigate the convergence of collaborative ML and blockchain, to address some existing issues of traditional ML model training. In this section, we explore different ways of the combination of these two technologies, describing their applications and innovations. We group the ways of combination into two categories:

Protocol that incentivizes data sharing. Data sharing facilitates continuous ML model training and more optimal model performance can be obtained. With a blockchain-based protocol (named smart contract), data contributors will get rewards after providing reliable data and do not need to fear data abuse or data leakage. So the parties who want to train ML models on certain problems can have access to the necessary data sets.

Marketplace that encourages model sharing. There are also individuals or organizations that have data but have no solutions to certain problems. A marketplace built on top of blockchain can be a great convenience for them since it encourages model sharing and helps match data with the right ML model. Moreover, some other research works on blockchain-based marketplaces that involve both data sharing and model sharing, and we put them in this category as well. In addition, certain decentralized smart contract system ensures the privacy of financial transactions.

### 3.1 Protocol that Incentivizes Data Sharing

**Sharing Updatable Model (SUM) on Blockchain [8]:** This framework focuses on the sharing of data from different participants to train a ML model based on smart contracts. In this work, the ML model is trained on the IMDB reviews dataset [26] for sentiment classification and can improve its performance continuously through using collaboratively-built datasets. A schema of SUM is shown in **Fig. 3**. One of the most innovative parts of this SUM on blockchain is its incentive mechanisms built in a smart contract. The incentive mechanisms are applied in order to encourage better data submission, filter wrong or ambiguous data, and maintain model accuracy.

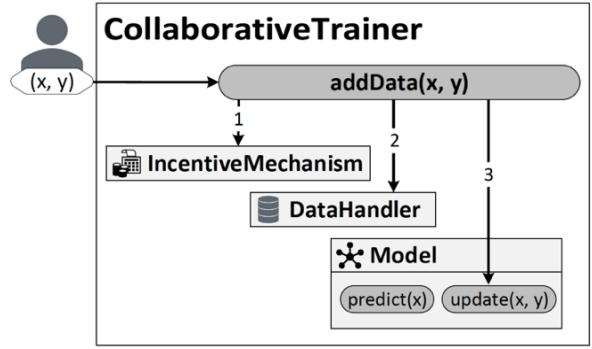

**Fig. 3.** The schema of SUM [8]: It requires 3 steps to add data to the model: (1) The IncentiveMechanism is responsible for validating the transactions. The developers give three ways of incentivization, including adding data with monetary or non-financial deposits. It may be triggered to provide users with rewards or punishment according to the uploaded data. (2) The DataHandler records all the related data to the blockchain to ensure accessibility for all future uses. (3) The ML model is updated according to training algorithms defined in advance. Any user can query the ML model for predictions as well as add their data to it.

The following are the three incentive mechanisms of the SUM framework:

*Gamification:* To reduce the barrier to entry, no-financial incentives such as points and badges can be awarded to reliable data contributors;

*Rewards mechanism based on prediction markets:* A monetary reward-based system is described for incentivizing the contribution of correct data. In this case, an outside party provides a pool of reward funds and a test dataset D, and the participants are rewarded according to how well they improve the performance of model h as measured by the test data. This monetary mechanism consists of three phases: Commitment Phase, Participation Phase, and Reward Phase. In the Commitment Phase, the provider deposits the reward, defines a loss function L (h, D) (limited to the range [0, 1] by the smart contract) for validation, uploads the test datasets, and specifies the end condition e.g. time limit. In the Participation Phase, the participants provide their dataset and the smart contract trains and updates the ML model. And in the Reward Phase, the smart contract updates the balance $b_t$ of each participant t through (1). The participants either get rewards or gain punishment on their deposit based on the quality of their data.

$$b_t = b_{t-1} + L(h_{t-1}, D) - L(h_t, D) \qquad (1)$$

In (1), $h_t$ represents the model updated by participant t. $b_{t-1}$ and $h_{t-1}$ respectively presents the balance of the previous participant and the model renewed by his/her dataset. The initial value of $b_t$ is 1 and all the participants whose balance is less than 1 after validation will be dropped out.

*Deposit, refund, and take: self-assessment:* To facilitate penalties on those submitting bad data, this proposal also enforces a deposit when contributing data. Therefore, it can reduce the negative influence of malicious agents by detaining their deposits of them. It relies on the data contributors to indirectly validate and pay each other without an outside party. This monetary mechanism consists of four phases: Deployment Phase, Deposit Phase, Refund Phase, and Validation Phase. In the Deployment Phase, a ML model h is trained in advance. In the Deposit Phase, the participants deposit some currency d each time they provide data x and label y. The deposit is influenced by the time interval between updates (2). After a time t passed, the participant whose data agrees with h(x) == y will have his/her entire deposit d returned. The relationship between the time t to wait for a refund and the probability of correctness of the model P(h(x) = y) is shown in (3). Since model h has been trained with a portion of data before Deposit Phase and has an initial accuracy, P cannot be zero in (3). In the Validation Phase, the smart contract has data validated, rewards the participants providing good data and takes a portion of deposit d of each participant whose validation result is h(x) ≠ y. **Fig. 4**. shows the result of the simulation experiment of the third incentive mechanism.

$$d \propto \frac{1}{time\ since\ last\ update} \qquad (2)$$

$$t \propto \frac{1}{P(h(x)=y)} \quad (t \geq 7 days) \qquad (3)$$

The SUM proposes a feasible and effective platform for participants to collaboratively build a dataset and uses smart contracts to host a continuously updated and stable ML model, according to the demo and simulation experiments done by the developers of this framework. The financial and non-financial incentive structures can provide good data to the models and remain their accuracy.

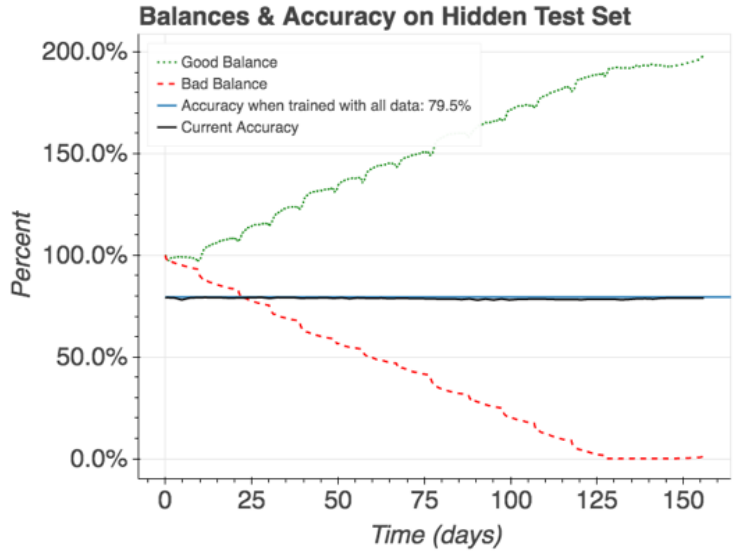

**Fig. 4.** The result of the simulation experiment of SUM [8]: The percentage of the balance and the accuracy of the ML model is shown in the figure. The Good Agent who provides good data gets rewards while the Bad Agent who uploads bad data runs out of the balance eventually. The smart contract filters out the negative influence of the bad data and ensures the stability of the model.

**Blockchain Empowered FL System I [11]:** The system focuses on the secure sharing of data generated from connected devices in the Industrial Internet of Things (IIoT) paradigm. It gives us a creative idea of data privacy protection, that is to share data models instead of revealing the actual data. The raw data owned by the contributors are stored locally by themselves, hence there is no need for fear of data leakage.

The proposed system in **Fig. 5**. consists of two main modules: the permissioned blockchain module and the FL module. The Permissioned blockchain in the middle establishes secure connections among all the end IoT devices through its encrypted records. It manages the accessibility of data and records retrieval transactions, data sharing actions, other related data, and all the sharing events of data, instead of recording raw data. The FL module at the bottom shares the federated data model learned over decentralized multiple parties. An originally created consensus is the Proof of Training Quality (PoQ) which decreases the computing cost as well as the communication resources.

Reflected from the evaluation results of the developers of this system, it is clear that the blockchain-empowered data sharing scheme can enhance the secure data sharing process. Moreover, by integrating FL into the consensus process of blockchain permissioned, the developers not only improved the utilization of computing resources but also increased the efficiency of the data sharing process.

Therefore, it is a promising way to combine FL with blockchain to ensure data privacy in the process of data sharing.

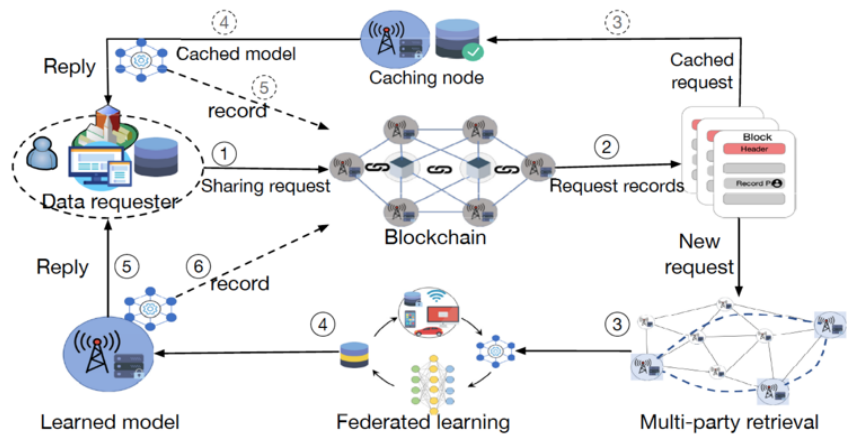

**Fig. 5.** The working mechanism with the architecture of secure data sharing [11]: The Data requester launches a data-sharing requirement to the permissioned blockchain (blockchain). Then the blockchain encrypts and records the information on the blocks. Multi-party trained the model with new data and the Data requester gains the federated data model which is also recorded to the blockchain. The Cached model is used to check whether the request has been processed before to avoid repetitive operations.

**Blockchain Empowered FL System II [27]:** This framework is also a combination of FL and blockchain that secures the share of data like the previous one. The difference is that it uses Proof of Work (PoW) instead of PoQ. The updates of the locally trained model are written in the transaction of the blockchain which ensures secured updates and are broadcasted to other nodes automatically. Then, through the validation of all the other nodes and the confirmation of work by the majority of nodes, the update is applied to the current model. Compared to traditional FL, this framework diminishes the issues of malicious servers and unbelievable data holders and preserves privacy.

**DeepChain [10]:** This is another way of ensuring data privacy while creating a protocol that incentivizes data sharing. According to the developers of the DeepChain prototype, the distributed, secure, and fair deep learning framework solves many security problems neglected in FL and provides a value-driven incentive mechanism based on blockchain to force the participants to behave correctly. A schema of this framework is shown in **Fig. 6**.

The developers also implemented the DeepChain prototype in their paper. They use decentralized ledger Corda V3.0 [27] and MNIST dataset [28] to build a blockchain for simulation experiments.

In the evaluation process, the model is trained on the DeepChain in a multi-party setting by using cipher size, throughput, training accuracy, and total cost of time and obtain results such as the more parties jointly train the model, the higher the training accuracy; the throughput decreases while there is a larger number of gradients; the time for training increased as the number of the parties grows.

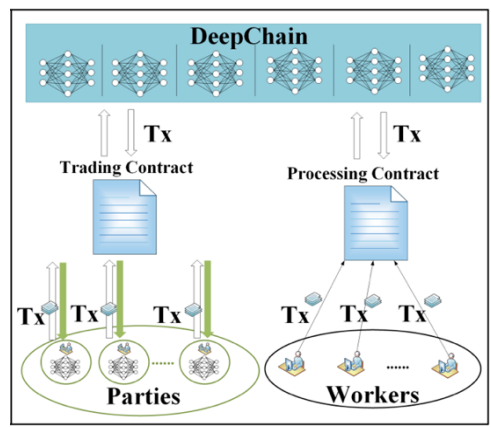

**Fig. 6.** DeepChain prototype [10]: Parties are the entities that have similar needs but are unable to independently train the model that meets their requirements. Trading Contract allows the Parties to upload their local gradients to the DeepChain. Workers are like miners in Bitcoin who record transaction data to the blockchain and gain rewards computed by Processing Contract. Trading Contract and Processing Contract are smart contracts in DeepChain, together guiding the secure training process, while Tx refers to the transaction.

The main contributions of DeepChain are: The DeepChain proposed with an incentive mechanism encourages joint participation in deep learning model training and sharing of the obtained local gradients; The DeepChain guarantees both the privacy of local gradients and the auditability of the training process.

Considering the methods mentioned above, the convergence of ML and blockchain is promising to empower the development of ML through encouraging continuous and reliable data sharing.

**LearningChain [30]:** The LearningChain considers a general (linear or nonlinear) learning model on a blockchain system. This Ethereum-based framework contains three main phases: The first phase, blockchain initialization, sets up a peer-to-peer network with nodes and data holders; The second phase, local gradient computation, provides copies of the model to the data holders who compute the gradients locally and add noise factor to them to ensure users' privacy; The third phase, global gradient aggregation, generates the winning node through a Proof of Work (PoW) problem and updates the model by the upgraded global gradient. Through the application of blockchain, there exists a mutual restraint between users' privacy and model accuracy in that the decrease of privacy leads to an increase in test errors. The comparison of the test errors between learning chains indicates that the use of blockchain in the Learning Chain is efficient and effective.

### 3.2 Marketplace that encourages model sharing

**DanKu Contract [12]:** This classic and original protocol is built on top of an Ethereum-blockchain-based marketplace for evaluating and exchanging ML model, which is also a trustless platform for supply and demand. The demander (organizer) who has the data set brings out a question, and the supplier (submitter) submits their well-trained ML model to solve this problem, and in the finalize stage, the one whose model performance ranks highest can have a payout from the demander. Hashed data is revealed through the contract to ensure data privacy and the blockchain is also utilized to record all transactions. A schema of this marketplace is shown from **Fig. 7**. to **Fig. 11**. These five stages ensure the successful execution of the marketplace. As far as we've concerned, there are several innovations in this marketplace:

*Automation and Anonymity:* The exchange of ML models in this marketplace is automated and anonymous, unlike the Ethereum blockchain, which requires identities or reputations. That is because the protocol enforces transactions through cryptographic verification.

*No Need for Escrow System:* The utilization of the smart contract, which uses the blockchain to automatically validate the solutions submitted by the submitters, eliminates the doubt of whether the solution is correct or not. On this basis, the organizers that are seeking an AI solution to a problem can solicit globally, while the submitters who are managed to solve the problem can directly exchange monetize rewards with their ML models.

*Data Privacy:* To ensure data privacy, the DanKu contract helps the organizer create a cryptographic dataset for training and testing by using the hashing function (sha3-keccak [29]).

The developers of this marketplace, A. B. Kurtulmus and K. Daniel points out in their paper that the smart contract creates a market where parties who are good at solving ML problems can directly monetize their skillset and where any organization or software agent that has a problem to solve with AI can solicit solutions from all over the world. This will incentivize the creation of better ML models, and make AI more accessible to companies and software agents [12]. That would be a great contribution to the development of ML.

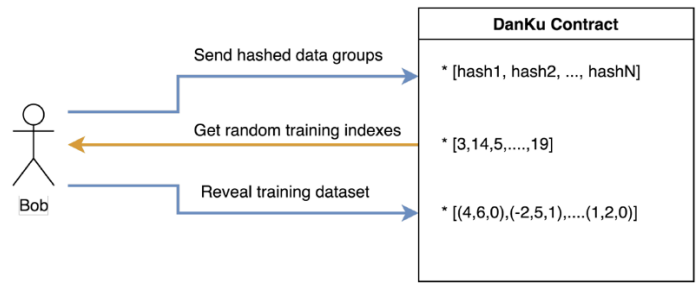

**Fig. 7.** Initialization Stage [12]: The organizer creates a contract with the problem they wish to solve (expressed as an evaluation function), datasets for testing and training, and an Ethereum wallet. Hashed training data sets are revealed.

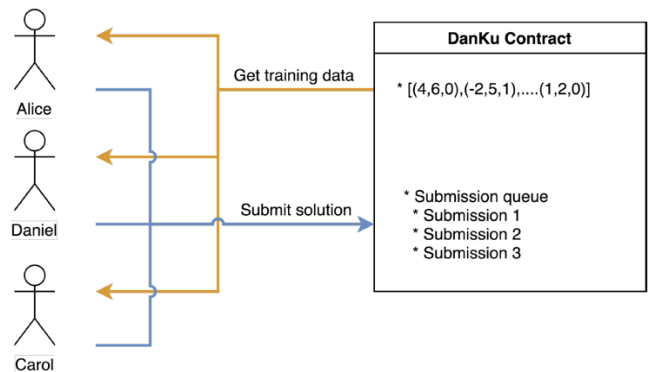

**Fig. 8.** Submission Stage [12]: Submitters provide their potential solutions (trained ML models).

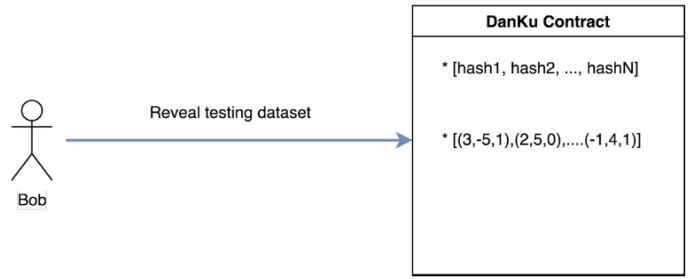

**Fig. 9.** Test Dataset Reveal Stage [12]: The organizer reveals the hashed testing data sets.

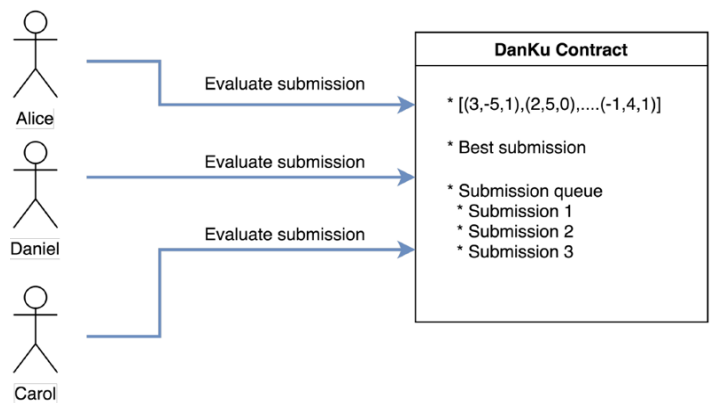

**Fig. 10.** Evaluation Stage [12]: The evaluation function is called to mark the submitted model that passes the evaluation and is better than the best model submitted so far (or is the first model ever evaluated).

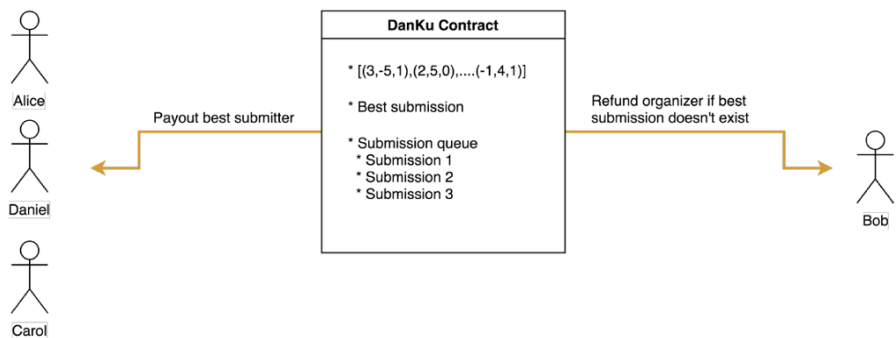

**Fig. 11.** Finalize Stage [12]: The best model submitter gets the reward, or the reward is paid back to the organizer if the best model does not exist (no model passes the evaluation stage).

**Proof-of-Learning (PoL) [33]:** This is also a cryptocurrency-based protocol that ranks the ML models for a given task and builds a distributed consensus system with incentive mechanisms. Three types of nodes participant in this framework: supplier nodes provide tasks previously; trainer nodes submit ML models; validator nodes rank the trained models in a distributed method according to their performance metric. The trainer node whose model performs best gets the reward, and the validators receive transaction fee paid by the supplier nodes.

PoL builds a symbiotic relationship between these two tasks: 1) validating transactions in a distributed ledger, and 2) storing ML models and experiments in a distributed database. It alleviates the computational waste involved in block mining and verifies the database and experiments automatically.

**DInEMMo [14]:** The full name of this architecture is Decentralized Incentivization for Enterprise Marketplace Models. It is a more complete marketplace for both model and data sharing and is conducted on the convergence of decentralized AI and blockchain. Users can upload a new ML model on this marketplace or enhance the existing ML models by submitting their datasets. A schema of this marketplace is shown in **Fig. 12**.

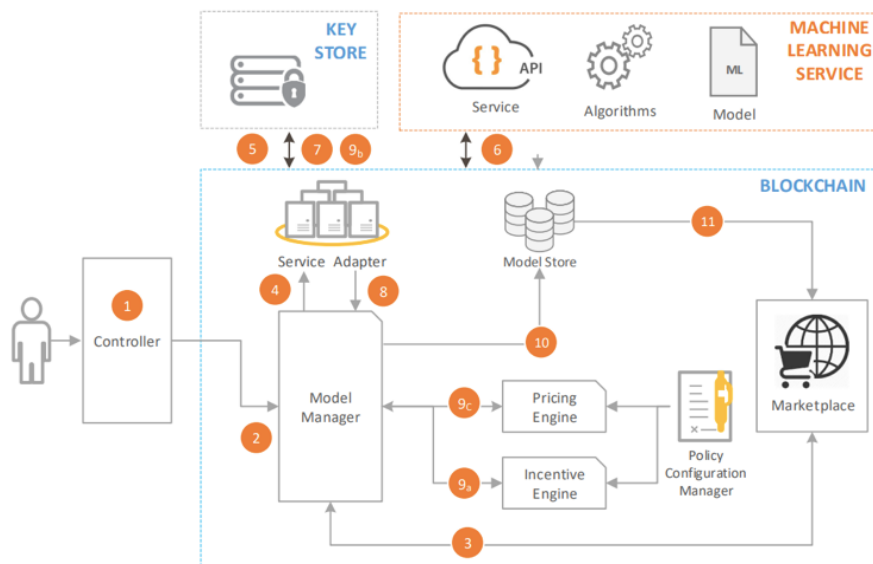

**Fig. 12.** DInEMMo Architecture [14]: DInEMMo is composed of a BLOCKCHAIN module, a KEY STORE module, and a MACHINE LEARNING SERVICE module. First, the Controller of the marketplace collect the dataset from the users and invokes the Model Manager, which interacts with the Marketplace and search for a suitable model through validation. Second, the Service Adapter communicates to the MACHINE LEARNING SERVICE module and obtains the key of the selected model that ensures model privacy from the KEY STORE module. Then, the Service Adapter updates the newly-trained ML model, and the Model Manager uploads it to the Model Store for further use. The Incentive Engine computes the rewards of data contributors and the Pricing Engine determines the model price.

As is explored in the paper, the applicability of the marketplace is shown by a use case scenario on medical diagnostic and the dataset of two hospitals consist of information [30] about various patients. In this scenario, the DInEMMo framework provides a marketplace for hospitals to share data and collaboratively enhance ML models without revealing patients' raw medical records. The price of the model depends on the recall, accuracy, and precision of the trained model.

In general, this architecture can realize the following functions:

First, the user can select the model that best fits the data he/she submits by using a validation mechanism in the marketplace;

Second, if the dataset submitted by the user helps the ML model perform better, a reward will be paid to the user according to the incentive mechanism (IE). This is quite different from the incentive method in DanKu contract, which only rewards the submitter whose model ranks first;

Third, the user seeking a solution to a ML problem can also buy a suitable model from the marketplace, with a pricing mechanism (PE) calculating the price for the models.

The developers of this marketplace put forward that the DInEMMo is the first-of-a-kind which fairly incentivizes the contributors of the ML models; based on the user policy, weigh the domain properties in addition to the ML attributes while rewarding the asset owners.

They also point out that DInEMMo is enabled with configurable smart contracts with the following features: represent the ML model and use case attributes; generate the ML models (new/enhanced) based on user input; compute the price of the ML model based on the user policy; calculate the incentives to the ML model's owner and co-contributors.

**Hawk [15]:** The main issue that Hawk focuses on is transactional privacy, which is neglected in many existing systems where all transactions data are exposed clearly and publicly on the blockchain. This lack of privacy on the transaction is a major hindrance towards the adoption of smart contracts in the fields such as financial services where transactions data are considered to be highly confidential. The overview of Hawk is shown in **Fig. 13**. A Hawk program contains two parts: A private portion that takes in parties' input data and currency units and is meant to protect the participants' data and the exchange of money. The flow of money and the transacted amount of money are hidden in this private portion by sending encrypted data to the blockchain; a public portion that does not touch data or money.

In [15], the authors developed this decentralized smart contract system that does not store financial transactions in the clear on the blockchain, thus retaining transactional privacy from the public's view. A Hawk programmer can write a private smart contract that defines rules that govern financial fairness without having to implement cryptography since the Hawk automatically generates a cryptographic protocol. Both the Hawk user and the manager (the monitor) can execute the Hawk contract. The manager does not need to be trusted since he/she will be penalized when abortions happened and the user will gain compensation. As defined by the authors, Hawk is the first to formalize the blockchain model of cryptography.

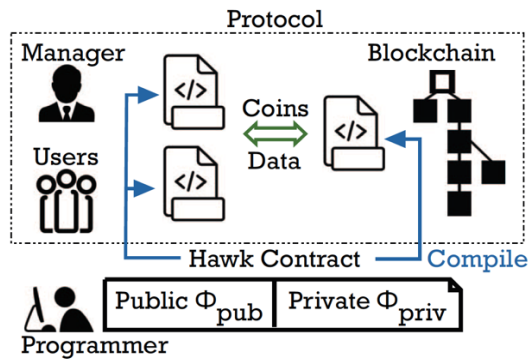

**Fig. 13.** Hawk overview [15]: The programmer defines the rules of a Hawk protocol that ensures financial fairness. Both the manager and the user can execute the protocol. The user can commit not only normal data but also coins to the protocol. The manager monitors the transaction but cannot affect any of its outcomes, and will get penalized if acting against honesty. The private portion cryptographically hides the money-related privacy data and the pubic portion contains structures such as the incentive mechanism that do not touch money or other private data.

## 4 Limitations and Future Research Directions

While the convergence of ML and blockchain has many advantages, there are still some challenges that may hinder its development and cannot be overlooked. In this section, we list some limitations of the current combination of ML and blockchain and some possible future research directions of these applications.

### 4.1 Incentive Mechanisms

Firstly, we discuss the incentive mechanisms, which are effective ways to encourage sharing of data and ML models. There may be a potential overfitting problem in the reward phase. For example, we think the problem may appear in the Rewards mechanism based on prediction markets of [8] because it may not be feasible to judge the validity of data solely on the decline of the lost function. That is because when overfitting occurs, a positive reward could not be proof of good data submission. And even though the model is well trained, the lost function here is not enough to describe or capture the model generalization capability. According to the authors of [8], more exploration, analysis, and experiments with incentive mechanisms in this space need to be done in the future with an emphasis on the type of model each incentive mechanism works well with, as well as the risk of overfitting. So, we propose the following possible improvements:

First, the developer of [8] gives us an idea to exclude data providers that fail to optimize the model to be rewarded through (1). However, we think the drop-out condition in it could be further fine-tuned. For instance, the smart contract can drop out the participant when his/her balance $b_t<1.1$ instead of $b_t<1$, or the balance $b_t<1$ happens twice in the Reward Phase;

Second, we believe that the form of the lost function could be optimized. For instance, we can control the number of training epochs that the lost function in (1) iterates by setting up a limit. In this case, a particularly small and keep-diminishing value of the lost function can be avoided and the risk of overfitting may be reduced compared with the existing approach in [8]. The reason is that we should concern about the improvement of the model performance on the validation dataset in addition to the decrease of the lost function. We wish a good performance of the trained model but not an extremely good one, such as a 99.99% accuracy with a risk of overfitting. Hence, we propose performing cross-validation [31] to the model to alleviate the problem of overfitting. By splitting the original dataset into the training set and validation set, cross-validation aims at testing the model's ability of prediction by using split data that was not used to train the model and provides unbiased evaluation. It also gives an insight into how the trained model will generalize to a new dataset;

Third, the Incentive Engine in [14] mainly considers the change of accuracy when computing the reward of contributors. Nevertheless, in the case where we have imbalanced data, accuracy is not a very representative model performance metric. So, we think it could be improved by using more comprehensive evaluation metrics such as F1-score, and considering additional factors like the size of a training data size, etc. F1-score is the harmonic means of Precision (the percentage of truly positive data in all the data that are predicted positive) and Recall (the percentage of the data that are predicted positive in all truly positive data).

### 4.2 Privacy

The importance of data privacy cannot be ignored. Contributors may think it unsafe to expose their data or transactions on the blockchain and not want to publish their data to a public blockchain through which every user with the key can have access to the data. Possible ways to solve this issue are listed below:

First, as proposed by the authors of [8], future work can be done to not submit data directly to the smart contract. Contributors can just submit encoded input or submit input to a hidden model, which hides behind an application program interface (API, an added layer) and is not necessarily public;

Second, blockchain technology, if effectively used, can also be helpful to standardize and track the utilization of recorded data, thus avoiding data abuse and privacy violations. Smart contracts can be optimized to reveal hashed data or data models instead of uploading raw data as is done in [11] [12];

Third, in [12], the developers also point out that the model weights are not completely anonymized. That is to say, any user can access the model weights submitted on the DanKu contracts. A possible way to solve this issue is to include homomorphic encryption in the protocol to anonymize the models submitted on the smart contracts;

Furthermore, many of the protocols or marketplaces so far do not consider transaction privacy, and so the further study may be necessary for conjunction with [15].

### 4.3 Cost Control

The organizers of the protocol and marketplace also need to take into account the control of the cost e.g. the gas fee. The amount of gas needed to store dataset, model, transaction, and other related data in the smart contract and the marketplace is significant. Complex models may not be able to run due to the limitation of gas for running smart contracts. Potential solutions given in [12] can help reduce gas costs. Store large files and datasets in alternatives like IPFS [32] and swarm instead of blockchain, so the keeping price would be at a reasonable level. The solidity language could introduce new features to make smart contracts faster and cheaper. Improvements in ML such as using 8-bit integers will also help reduce the cost. Most popular ML libraries might need to be adapted to work with integers instead of floating points since solidity only works with integers.

### 4.4 Complex Models

Restricted to storage space and computing resources of blockchain, many existing architectures of the combination of ML and blockchain are mainly designed with basic simple models e.g. perceptron model. However, the solution to more complex problems requires more complex models. In certain applications in more complicated scenarios such as traffic signs detection in autonomous driving, the detecting results are affected by the weather, lights, angles, etc., and involves huge amounts of collaboratively collected data. Therefore, these systems can be efficiently updated by designing for the utilization of more complex models e.g. deep learning models to learn more sophisticated patterns from the data, solve more complicated issues, and realize continuous training.

## 5 Conclusions

In this paper, we provide an investigation of the convergence of ML and blockchain. Firstly, a brief introduction of ML and blockchain is given respectively. Then we detail the applications and innovations of this combination, the ways of which are grouped into two categories: protocol that incentivizes data sharing and marketplace that encourages model sharing. Finally, analyses are given on the limitations of existing research on incentive mechanisms, data and transaction privacy, cost control on protocol and marketplace design, and the utilization of complex models. Future research directions are also provided to the improvement of these limitations. From this survey, we can conclude that compared to the traditional ML, the integration of ML and blockchain allows for effective sharing of data and model; By using the blockchain technology, model training can be done even if there is malicious data in the dataset; Continuous new data makes the trained model more stable and more accurate; It is also easier to screen the data and eliminate the influence of bad data; At the same time, data security and privacy issues can be guaranteed. By leveraging advances in ML and blockchain, the new kind of framework is sure to have a promising future to the development of fields of ML and blockchain.